\documentclass[acmtog,screen,nonacm]{acmart}

\usepackage{booktabs} % For formal tables
\usepackage{amsfonts}
\usepackage{pdfpages}
\usepackage{soul}

\usepackage[ruled]{algorithm2e} % For algorithms

\SetAlFnt{\small}
\SetAlCapFnt{\small}
\SetAlCapNameFnt{\small}

\SetAlCapHSkip{0pt}
\usepackage{enumitem}

\begin{document}

\newcommand{\targetimg}{\mathbf{T}}  
\newcommand{\tindex}{j} 
\newcommand{\pt}{x} 
\newcommand{\viewdir}{v} 
\newcommand{\handpose}{p}  
\newcommand{\signeddist}{d}  
\newcommand{\featureop}{f_{32}} 

\newcommand{\etal}{\textit{et al.}}

\def\reta{{\textnormal{$\eta$}}}
\def\ra{{\textnormal{a}}}
\def\rb{{\textnormal{b}}}
\def\rc{{\textnormal{c}}}
\def\rd{{\textnormal{d}}}
\def\re{{\textnormal{e}}}
\def\rf{{\textnormal{f}}}
\def\rg{{\textnormal{g}}}
\def\rh{{\textnormal{h}}}
\def\ri{{\textnormal{i}}}
\def\rj{{\textnormal{j}}}
\def\rk{{\textnormal{k}}}
\def\rl{{\textnormal{l}}}
\def\rn{{\textnormal{n}}}
\def\ro{{\textnormal{o}}}
\def\rp{{\textnormal{p}}}
\def\rq{{\textnormal{q}}}
\def\rr{{\textnormal{r}}}
\def\rs{{\textnormal{s}}}
\def\rt{{\textnormal{t}}}
\def\ru{{\textnormal{u}}}
\def\rv{{\textnormal{v}}}
\def\rw{{\textnormal{w}}}
\def\rx{{\textnormal{x}}}
\def\ry{{\textnormal{y}}}
\def\rz{{\textnormal{z}}}

\def\rvepsilon{{\mathbf{\epsilon}}}
\def\rvtheta{{\mathbf{\theta}}}
\def\rva{{\mathbf{a}}}
\def\rvb{{\mathbf{b}}}
\def\rvc{{\mathbf{c}}}
\def\rvd{{\mathbf{d}}}
\def\rve{{\mathbf{e}}}
\def\rvf{{\mathbf{f}}}
\def\rvg{{\mathbf{g}}}
\def\rvh{{\mathbf{h}}}
\def\rvu{{\mathbf{i}}}
\def\rvj{{\mathbf{j}}}
\def\rvk{{\mathbf{k}}}
\def\rvl{{\mathbf{l}}}
\def\rvm{{\mathbf{m}}}
\def\rvn{{\mathbf{n}}}
\def\rvo{{\mathbf{o}}}
\def\rvp{{\mathbf{p}}}
\def\rvq{{\mathbf{q}}}
\def\rvr{{\mathbf{r}}}
\def\rvs{{\mathbf{s}}}
\def\rvt{{\mathbf{t}}}
\def\rvu{{\mathbf{u}}}
\def\rvv{{\mathbf{v}}}
\def\rvw{{\mathbf{w}}}
\def\rvx{{\mathbf{x}}}
\def\rvy{{\mathbf{y}}}
\def\rvz{{\mathbf{z}}}

\def\erva{{\textnormal{a}}}
\def\ervb{{\textnormal{b}}}
\def\ervc{{\textnormal{c}}}
\def\ervd{{\textnormal{d}}}
\def\erve{{\textnormal{e}}}
\def\ervf{{\textnormal{f}}}
\def\ervg{{\textnormal{g}}}
\def\ervh{{\textnormal{h}}}
\def\ervi{{\textnormal{i}}}
\def\ervj{{\textnormal{j}}}
\def\ervk{{\textnormal{k}}}
\def\ervl{{\textnormal{l}}}
\def\ervm{{\textnormal{m}}}
\def\ervn{{\textnormal{n}}}
\def\ervo{{\textnormal{o}}}
\def\ervp{{\textnormal{p}}}
\def\ervq{{\textnormal{q}}}
\def\ervr{{\textnormal{r}}}
\def\ervs{{\textnormal{s}}}
\def\ervt{{\textnormal{t}}}
\def\ervu{{\textnormal{u}}}
\def\ervv{{\textnormal{v}}}
\def\ervw{{\textnormal{w}}}
\def\ervx{{\textnormal{x}}}
\def\ervy{{\textnormal{y}}}
\def\ervz{{\textnormal{z}}}

\def\rmA{{\mathbf{A}}}
\def\rmB{{\mathbf{B}}}
\def\rmC{{\mathbf{C}}}
\def\rmD{{\mathbf{D}}}
\def\rmE{{\mathbf{E}}}
\def\rmF{{\mathbf{F}}}
\def\rmG{{\mathbf{G}}}
\def\rmH{{\mathbf{H}}}
\def\rmI{{\mathbf{I}}}
\def\rmJ{{\mathbf{J}}}
\def\rmK{{\mathbf{K}}}
\def\rmL{{\mathbf{L}}}
\def\rmM{{\mathbf{M}}}
\def\rmN{{\mathbf{N}}}
\def\rmO{{\mathbf{O}}}
\def\rmP{{\mathbf{P}}}
\def\rmQ{{\mathbf{Q}}}
\def\rmR{{\mathbf{R}}}
\def\rmS{{\mathbf{S}}}
\def\rmT{{\mathbf{T}}}
\def\rmU{{\mathbf{U}}}
\def\rmV{{\mathbf{V}}}
\def\rmW{{\mathbf{W}}}
\def\rmX{{\mathbf{X}}}
\def\rmY{{\mathbf{Y}}}
\def\rmZ{{\mathbf{Z}}}

\def\ermA{{\textnormal{A}}}
\def\ermB{{\textnormal{B}}}
\def\ermC{{\textnormal{C}}}
\def\ermD{{\textnormal{D}}}
\def\ermE{{\textnormal{E}}}
\def\ermF{{\textnormal{F}}}
\def\ermG{{\textnormal{G}}}
\def\ermH{{\textnormal{H}}}
\def\ermI{{\textnormal{I}}}
\def\ermJ{{\textnormal{J}}}
\def\ermK{{\textnormal{K}}}
\def\ermL{{\textnormal{L}}}
\def\ermM{{\textnormal{M}}}
\def\ermN{{\textnormal{N}}}
\def\ermO{{\textnormal{O}}}
\def\ermP{{\textnormal{P}}}
\def\ermQ{{\textnormal{Q}}}
\def\ermR{{\textnormal{R}}}
\def\ermS{{\textnormal{S}}}
\def\ermT{{\textnormal{T}}}
\def\ermU{{\textnormal{U}}}
\def\ermV{{\textnormal{V}}}
\def\ermW{{\textnormal{W}}}
\def\ermX{{\textnormal{X}}}
\def\ermY{{\textnormal{Y}}}
\def\ermZ{{\textnormal{Z}}}

\def\vzero{{\bm{0}}}
\def\vone{{\bm{1}}}
\def\vmu{{\bm{\mu}}}
\def\vtheta{{\bm{\theta}}}
\def\va{{\bm{a}}}
\def\vb{{\bm{b}}}
\def\vc{{\bm{c}}}
\def\vd{{\bm{d}}}
\def\ve{{\bm{e}}}
\def\vf{{\bm{f}}}
\def\vg{{\bm{g}}}
\def\vh{{\bm{h}}}
\def\vi{{\bm{i}}}
\def\vj{{\bm{j}}}
\def\vk{{\bm{k}}}
\def\vl{{\bm{l}}}
\def\vm{{\bm{m}}}
\def\vn{{\bm{n}}}
\def\vo{{\bm{o}}}
\def\vp{{\bm{p}}}
\def\vq{{\bm{q}}}
\def\vr{{\bm{r}}}
\def\vs{{\bm{s}}}
\def\vt{{\bm{t}}}
\def\vu{{\bm{u}}}
\def\vv{{\bm{v}}}
\def\vw{{\bm{w}}}
\def\vx{{\bm{x}}}
\def\vy{{\bm{y}}}
\def\vz{{\bm{z}}}

%\title{FPH: Fast and Photorealistic Dynamic Head Rendering via 3D Gaussians}
\title{GaussianHeads: End-to-End Learning of Drivable Gaussian Head Avatars from Coarse-to-fine Representations}

%% Authors
\author{Kartik Teotia}
\email{kteotia@mpi-inf.mpg.de}

\affiliation{%
  \institution{Max Planck Institute for Informatics and Saarland Informatics Campus}
  \country{Germany}
}

\author{Hyeongwoo Kim}
\email{hyeongwoo.kim@imperial.ac.uk}

\affiliation{%
  \institution{Imperial College London}
  \country{United Kingdom}
}

\author{Pablo Garrido}
\email{pablo.garrido@flawlessai.com}

\affiliation{%
  \institution{Flawless AI}
  \country{United States}
}

\author{Marc Habermann}
\email{mhaberma@mpi-inf.mpg.de}

\affiliation{%
  \institution{Max Planck Institute for Informatics and Saarland Informatics Campus}
  \country{Germany}
}

\author{Mohamed Elgharib}
\email{elgharib@mpi-inf.mpg.de}

\affiliation{%
  \institution{Max Planck Institute for Informatics}
  \country{Germany}
}

\author{Christian Theobalt}
\email{theobalt@mpi-inf.mpg.de}

\affiliation{%
  \institution{Max Planck Institute for Informatics and Saarland Informatics Campus}
  \country{Germany}
}

\begin{teaserfigure}
\centering
  \includegraphics[width=1\textwidth]{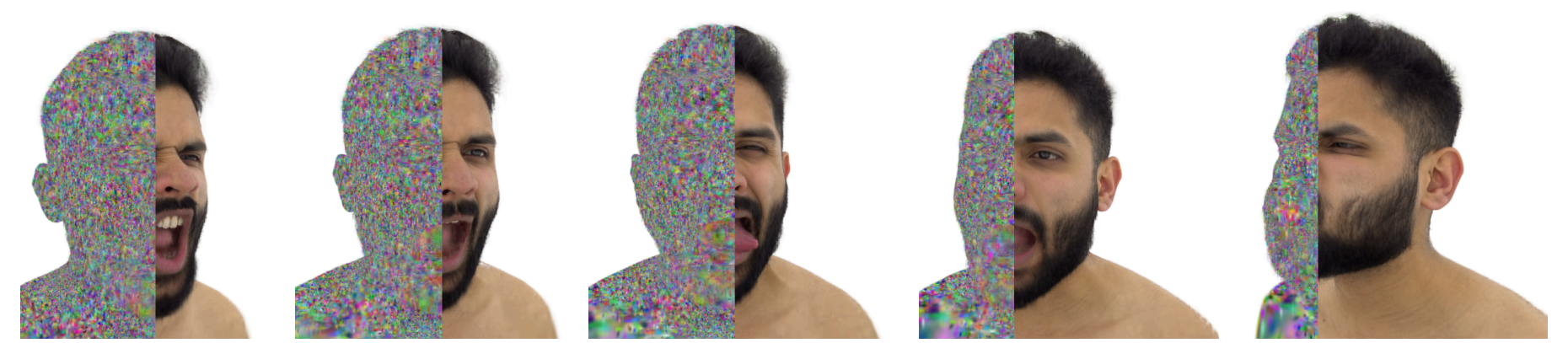}
  %\caption{Our method generates highly photorealistic renderings of human heads in motion at a rendering speed of $100$ for $960x540$ resolution. Our method is driven by a monocular RGB image, and leverages 3D Gaussians as the underlying primitive representation to model the dynamics of the human head. We show results for both high-quality dynamic human head reconstruction as well controllable volumetric head renderings.}
  \caption{\textbf{GaussianHeads} renders photorealistic dynamic 3D human heads in real time. In particular, it renders fine-scale details of (facial) hair and large geometric deformations in the mouth region at high fidelity. 
  Each 3D Gaussian primitive, which is consistently tracked against large deformations over time, is visualized in colors.
  By providing input images as a control signal, our model can also be used for creating controllable head animations.
  %\HK{if time allows, provide zoom-in over tongue hairs etc, change `rendering' to `Rendering'}
  %\HK{rather than describing the application as being controllable, maybe better to use self-reenactment or something similar for clarity in the paper}
  %\ME{pending: teaser figure should have white background and get referenced}
  %\HK{update the figure?}
  }
  \label{fig:teaser}
\end{teaserfigure}

\begin{abstract}
   
Real-time rendering of human head avatars is a cornerstone of many computer graphics applications, such as augmented reality, video games, and films, to name a few.
Recent approaches address this challenge with computationally efficient geometry primitives in a carefully calibrated multi-view setup.
Albeit producing photorealistic head renderings, they often fail to represent complex motion changes, such as the mouth interior and strongly varying head poses.
We propose a new method to generate highly dynamic and deformable human head avatars from multi-view imagery in real time.
At the core of our method is a hierarchical representation of head models that can capture the complex dynamics of facial expressions and head movements.
First, with rich facial features extracted from raw input frames, we learn to deform the coarse facial geometry of the template mesh.
We then initialize 3D Gaussians on the deformed surface and refine their positions in a fine step. 
We train this coarse-to-fine facial avatar model along with the head pose as learnable parameters in an end-to-end framework.
This enables not only controllable facial animation via video inputs but also high-fidelity novel view synthesis of challenging facial expressions, such as tongue deformations and fine-grained teeth structure under large motion changes.
Moreover, it encourages the learned head avatar to generalize towards new facial expressions and head poses at inference time.
We demonstrate the performance of our method with comparisons against the related methods on different datasets, spanning challenging facial expression sequences across multiple identities. We also show the potential application of our approach by demonstrating a cross-identity facial performance transfer application. We make the code available on our \href{https://vcai.mpi-inf.mpg.de/projects/GaussianHeads/}{\textbf{project page}}.

%%%%%%%%%%%%%%%%%%%%%%%%%
%
 
\end{abstract}


\maketitle

\section{Introduction} \label{sec:introduction}
 Photorealistic modeling and rendering of human heads is essential in applications such as virtual telepresence, video games, and movies. Achieving an immersive experience in these applications is a long-standing research problem
 as it necessitates representing facial expressions with a high degree of detail while also ensuring real-time performance.
 Current approaches struggle with a trade-off: they either excel at rendering expressions in real time but with limited details in regions, such as mouth interior and hair~\cite{lombardi2018_deep-appear,ma2021_pixel-codec,lombardi21_mvp}, or offer highly detailed reconstructions at the expense of slower rendering speeds~\cite{NPVA,kirschstein2023nersemble,teotia2023hq3davatar,gao2022_nerfblendshape}. 
 
Photorealistic avatars are commonly built on top of explicit mesh-based representations as they offer consistent topological changes and real-time rendering capabilities \cite{lombardi2018_deep-appear, ma2021_pixel-codec}. 
However, mesh-based representations present limited capacity to model facial structures exhibiting fine details like hair and beards \cite{teotia2023hq3davatar}. 
To overcome limitations commonly faced by explicit representations, Mixture of Volumetric Primitives (MVP)~\cite{lombardi21_mvp} proposes a volumetric representation based on unstructured cubic primitives. 
However, cube-based primitives have limited capability to represent fine structures~\cite{teotia2023hq3davatar, NPVA}.
HQ3DAvatar \cite{teotia2023hq3davatar} proposes an implicitly learned canonical representation and shows the capability of rendering HD images at interactive speeds, yet it suffers from blurring in regions undergoing large deformations. 
In this work, we propose a novel end-to-end framework for modeling and rendering dynamic human heads with a high level of photorealism in real time, as shown in Fig.~\ref{fig:teaser}.
Our method leverages a deformable template face mesh, learnable global transformation parameters as part of our end-to-end learning framework, and 3D Gaussian primitives~\cite{3DGS} in a coarse-to-fine strategy to represent the human head dynamics. 
This enables high-quality representation of facial details at real-time rates (75 FPS) on a resolution of $960\times540$.  
Our method is trained using multi-view images in a person-specific manner. During inference, our method takes the driving videos as input to control the digital avatar under novel expressions, global rigid head poses, and camera viewpoints. 
It enables controllability via an animation code and global rigid pose parameters, both parameterized by the driving RGB videos.  
Such parameterization drives the 3D Gaussians in the world space, together with a computationally efficient implementation of the Gaussian primitive properties, such as color and opacity predictions. 
Our final renderings are obtained via a tile-based rasterizer of the Gaussian primitives~\cite{3DGS}. Fig. \ref{fig:teaser} shows the deforming Gaussians overlaid onto the RGB renderings of an expression sequence.

More specifically, our method first estimates the global rigid head pose and other facial deformations via an image-based encoder. It then deforms and poses a template mesh in the coarse reconstruction step, which becomes covered with 3D Gaussian primitives in the following refinement step.
These primitives are allowed to move beyond regions not represented by the deformed template mesh via a learnable position decoder, and their appearance is learned via color and opacity decoders.
Our model is learned from multi-view RGB video using photometric and perceptual loss terms, and newly introduced geometric and temporal constraints in an end-to-end manner. At test time, only a feed-forward pass is performed through the learned network, significantly contributing towards the fast rendering speed.

In summary, we make the following contributions:
\begin{itemize}
\item We present a novel method that leverages 3D Gaussian primitives to generate volumetric head avatars. Our method is trained using multi-view RGB video streams via several losses, and during test time, it can render the moving head with large deformations at real-time rates.
\item We introduce a coarse-to-fine representation for head and deformation models consisting of a deformable mesh and Gaussian primitives to learn coarse and fine-grained geometry and appearance. This effectively improves the convergence of our model leading to significantly improved photorealism, especially in highly deforming regions, such as the mouth interior.
\item We propose an end-to-end training pipeline allowing us to jointly supervise the head pose, geometry, and appearance in a fully differentiable manner.
\end{itemize}
We evaluate our approach visually and numerically against ground truth data. Here, we ablate our method with different design choices to illustrate their importance in the overall performance. Our approach outperforms existing state-of-the art methods~\cite{lombardi21_mvp,teotia2023hq3davatar,qian24_gaussian-avatars} qualitatively and quantitatively. 

\section{Related Work} 
\label{sec:related_work}

\subsection{Scene Representations}
In recent years, scene representations with neural components have unlocked potential applications in 3D reconstruction. 
NeRF~\cite{mildenhall2022_nerf} and Deep-SDF~\cite{park2019_deepsdf} are early pioneering works in modeling 3D scene properties using an MLP-based implicit learning platform. 
While these methods demonstrated strong results, they suffer from network capacity issues due to inefficient scene encodings. 
Grid-based strategies such as Instant-NGP~\cite{mueller2022_instant-ngp} and other volumetric approaches~\cite{fridovich2022_plenoxels,lombardi2019_neural-volumes} have been proposed for more efficient 3D scene modeling. 
Such advances led to increased rendering quality as well as better runtime performance. 
Neuro-explicit representations use point-based priors to improve reconstruction quality and runtime performance further. For instance, Point-NeRF~\cite{xu2022_pointnerf} introduced a hybrid representation with point clouds and neural features to model a radiance field. Mixture of Volumetric Primitives (MVP)~\cite{lombardi21_mvp} used cubic primitives weakly attached to the tracked facial geometry to model dynamic 3D scene content.

3D Gaussian Splatting~\cite{3DGS} presented Gaussian primitives
%,initialized on pre-computed point cloud geometry,
as the base representation to model 3D scene content, with recent extensions allowing joint camera and 3D scene modeling~\cite{fu24_colmap-free}. Such advances have produced state-of-the-art static 3D scene reconstruction in terms of fidelity and runtime performance. Dynamic 3D Gaussians~\cite{luiten2023dynamic} extended the 3D Gaussian Splatting framework to dynamic scenes but lacked user control. 
Hence, the scene can only be replayed back from novel camera viewpoints (i.e., no new object deformations) at inference time. 
Our approach introduces a controllable 3D Gaussian splatting framework for dynamic human heads. It builds on the runtime capabilities and reconstruction fidelity introduced in 3D Gaussian Splatting. Furthermore, it can render the examined head from a novel camera viewpoint under different expressions at test time. 
\vspace{-0.5em} % Adjust the value to reduce the 
\subsection{Mesh-based Head Avatars}
Several avatar generation methods rely on explicit scene-tracking provided by 3D morphable models (3DMMs) to aid with the modeling of dynamic expression changes. 
While there are some methods that operate purely in the 2D space~\cite{Recycle-GAN,Siarohin_2019_NeurIPS}, these methods often suffer from limited photorealism and generate talking faces with poor lip-sync. 
On the other hand, mesh-based scene representations have been used for generating personalized photorealistic head avatars of humans recorded in supervised multi-view studio setups~\cite{lombardi2018_deep-appear}. 
They have also been used in neural rendering pipelines to generate portrait avatars from 2D RGB data~\cite{thies2016face, tewari2018_self-supervised,kim2018_dvp,thies2019_dnr,wang2023styleavatar}.
%Mesh-based representations are limited by the underlying template fidelity, and thus, they usually struggle to represent fine details, such as scalp hair, beard, and mouth interior. 
However, the underlying template fidelity of mesh-based representations is limited. Thus, such representations usually struggle to represent fine details, such as scalp hair, beard, and mouth interior.

\begin{figure*}[htb]
\centering
\includegraphics[width=1\textwidth]{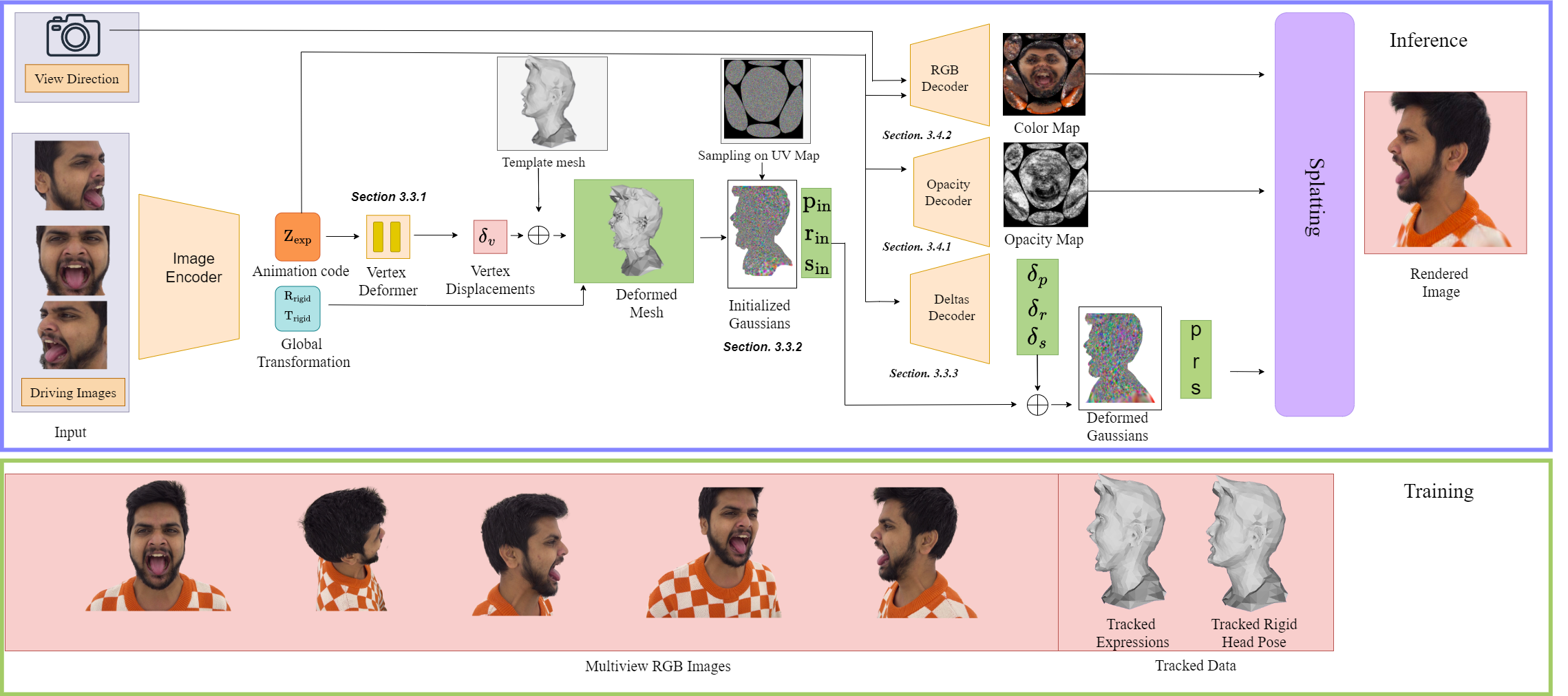}
\caption{\textbf{Method Overview}. Multi-view driving images are provided as input, and an image encoder extracts the animation code (\(\mathbf{Z}_{\text{exp}}\)) and global head pose parameters (\(\mathbf{R}_{\text{rigid}}\,\mathbf{T}_{\text{rigid}}\)). The animation code is fed into a vertex deformer network to generate per-vertex displacements (\(\delta_v\)) relative to a canonical template mesh in a rest pose with vertices $v_{t}$, resulting in an expression-dependent deformed mesh with vertices $v_{d}$. This is then globally transformed to a posed mesh with vertices $v_{p}$ using the head pose parameters (\(\mathbf{R}_{\text{rigid}}\,\mathbf{T}_{\text{rigid}}\)). 3D Gaussians with positions (\(\mathbf{p}_{\text{in}}\)), scales (\(\mathbf{s}_{\text{in}}\)), and rotations (\(\mathbf{r}_{\text{in}}\)) are initialized on the globally transformed deformed mesh. The deltas decoder predicts deformation for position (\(\delta_{p}\)), rotation (\(\delta_{r}\)), and scale (\(\delta_{s}\)) to refine the initialized 3D Gaussians. We employ two decoders to predict per-Gaussian RGB color and opacity aligned to the UV map of the template. 3D Gaussian Splatting projects the deformed Gaussians into the image plane, resulting in the rendered image. The pipeline is trained end-to-end using multi-view RGB images, expression tracking data, and rigid head pose tracking data. During testing, only a feedforward pass of the input is required to drive the global rigid head pose and facial expressions.
}

\label{fig:method}
\end{figure*}
%\vspace{-0.5em} % Adjust the value to reduce the 
\subsection{Volumetric Head Avatars}
Volumetric scene representations allow for capturing regions that are difficult to model using parametric mesh-based head trackers, e.g., the mouth interior. 
Hence, they have been an underpinning representation for human head avatars in recent years. Furthermore, their self-supervised training framework from 2D images makes them a very appealing approach.
In this section, we further review template-free as well as template-based approaches for volumetric human head avatars. 
\subsubsection{Template-based Approaches}
Most template-based approaches make use of a template mesh-guided canonical space to model facial expressions~\cite{athar2022rignerf, Zielonka2022InstantVH, gao2022_nerfblendshape, xu2023avatarmav,zheng2022imavatar, zheng23_point-avatar, zhao24_havatar} in a NeRF framework. Here, methods like INSTA~\cite{Zielonka2022InstantVH}, IM Avatar~\cite{zheng2022imavatar}, HAvatar~\cite{zhao24_havatar}, and PointAvatars\cite{zheng23_point-avatar} warp a world-space deformation field or points to canonical space using a nearest neighbor strategy to be able to deform points beyond the template representation. In contrast, our approach predicts per-frame tracking and learns to move Gaussians to regions not covered by the template in an end-to-end fashion.
Other approaches, such as HeadNeRF~{\cite{hong2021_headnerf} and that of Gafni~\etal~\cite{gafni2021_dynamic-nerf}} use 3DMM expression coefficients for conditioning the dynamic NeRF volume. 
Template-based NeRF avatars show impressive reenactment results but with limiting runtime performance due to the underlying coordinate-based representation. Mixture of Volumetric Primitives (MVP)~\cite{lombardi21_mvp} is a hybrid cube-based primitive model attached to a base mesh that renders head avatars in real-time at 1k resolution.
TRAvatar~\cite{yang23_travatar} builds upon MVP and uses a multispectral Lighstage to reconstruct animatable relightable avatars.
Unlike previous multi-view volumetric approaches~\cite{yang23_travatar, lombardi21_mvp}, our 3D Gaussian-based approach works with sparser camera rigs and simpler lighting setups. It also renders dynamic heads much faster thanks to its rasterization-based rendering approach and several important design choices. 
Such choices include a novel efficient vertex decoder and an efficient decoder-only architecture for predicting the properties of Gaussian primitives.
Due to the efficiency and rendering quality of 3D Gaussian Splats, research has quickly shifted its attention back to more explicit point representations, modeled as 3D displacements in the mesh surface space~\cite{qian24_gaussian-avatars, ma24_3d-gaussian-blendshapes, xu24_gaussian-head-avatar, rivero24_rig3dgs} or UV mesh space~\cite{Pang_2024_CVPR, saito24_relightable-gaussian, lan23_gaussian-3diff} of a base mesh or parametric model (e.g., FLAME~\cite{feng2021_deca}). Such representations allow for dynamic facial control via latent codes~\cite{saito24_relightable-gaussian} or expression parameters~\cite{qian24_gaussian-avatars}, while facial appearance can be conditioned on text-prompts~\cite{yang24_headstudio}.
Unlike concurrent multi-view work~\cite{saito24_relightable-gaussian,xu24_gaussian-head-avatar,qian24_gaussian-avatars}, we propose an end-to-end, coarse-to-fine deformation strategy that allows us to better estimate rigid mesh alignment, which in turn improves the alignment of Gaussians over the head surface, resulting in higher-quality renderings.

\subsubsection{Template-free Approaches} 
Template-free approaches are not limited by the representation capacity of the underlying 3DMM template. 
Here, methods such as Nerfies~\cite{park2021_nerfies} and HyperNeRF~\cite{park2021_hypernerf} use an implicitly learned canonical space to model dynamic scenes captured from a moving monocular camera. LatentAvatar~\cite{xu23_latent-avatar} learns implicit latent expression codes between shared and person-specific settings for cross-actor reenactment applications, thus improving tracking and expression transfer.
NeRFSemble~\cite{kirschstein2023nersemble} propose a multi-view solution for reconstructing dynamic human heads using a learned blending of multi-resolution hash grids.
While the methods above produce good results, their formulation does not extend beyond scene-replay, thus lacking controllability. 
GAN-based methods~\cite{chan2022_efficient-geom-aware} have shown improvements in reconstruction quality and view consistency. Thus, they have been used in downstream avatar-based applications~\cite{trevithick2023}.

HQ3DAvatar~\cite{teotia2023hq3davatar} is a more recent method that conditions a canonical space on expression features for photorealistic moving face synthesis.
The method is constrained using optical flow in a hash-grid radiance field framework~\cite{mueller2022_instant-ngp}. HQ3DAvatar produces high-fidelity reconstruction results and shows clear improvements over several existing methods, including MVP~\cite{lombardi21_mvp} and other multi-view extensions, such as HyperNeRF~\cite{park2021_hypernerf} and NeRFBlendShape~\cite{gao2022_nerfblendshape}. 
However, HQ3DAvatar struggles to reproduce details in regions undergoing strong deformations. Our approach benefits from the dense template mesh as a prior, yet it can flexibly deviate away from regions not covered by the underlying template. Such trade-off results in highly accurate renderings and real-time inference speeds, as demonstrated by extensive experiments.

\section{Method} \label{sec:method}
Our goal is to produce photorealistic, 3D-consistent moving heads at real-time rendering speeds. We use 3D Gaussians~\cite{3DGS} as the base representation, and introduce several novel losses and design choices to ensure fast rendering and high-quality reconstructions.
Our method, illustrated in Fig.~\ref{fig:method}, utilizes data from a subject's multi-view facial performance, captured with $24$ cameras sparsely distributed around the entire head. Such data supervises our end-to-end learning framework, which employs a coarse-to-fine strategy to capture head movements and detailed facial expressions accurately. To train our method, we track FLAME \cite{li2017_flame} parameters for each frame using a multi-view facial landmarks-based tracking implementation from \citet{DecafTOG2023}. At test time, our method requires a forward pass through the trained encoders and decoders to render the subject. 
Our method has an image encoder that separates input images into a local animation code and global transformations. The animation code drives deformations on top of a template mesh, which is then posed using the global transformation parameters. 3D Gaussians initialized on this posed mesh are refined to capture fine-scale details. The same animation code also drives the learning of opacity and RGB values. The Gaussian properties, as well as the RGB and opacity values, are learned in the 2D UV space of the template mesh, enabling the use of efficient CNN-based decoders.
This, combined with the fast rasterization of 3D Gaussian Splatting \cite{3DGS}, enables real-time inference. Notably, our method requires no explicit global rigid pose as an additional input, just the driving images during testing. Thus, it is more robust to jitter, which is usually introduced by explicit global rigid pose-tracking. In this section, we first lay out the image formation process via 3D Gaussian Splatting (Sec. \ref{preliminary}). Then, we describe our method in more detail, including our encoding strategy (Sec. \ref{sec:neuraltex}), our coarse-to-fine learning scheme (Sec. \ref{sec:c2f}), the proposed efficient decoders (Sec. \ref{sec:rgbadecoders}), and finally the objective functions (Sec. \ref{subsec:obj_functions}).

\subsection{Preliminaries: 3D Gaussian Splatting} \label{preliminary}
We adopt the image formation model from 3D Gaussian Splatting~\cite{3DGS}, which represents the scene using 3D Gaussian primitives. Each Gaussian primitive is represented by a $3\!\times\!3$ covariance matrix $\boldsymbol{\Sigma}$ centered at mean position $\mathbf{p}$ as follows:
\begin{equation}
g(\mathbf{x}) = e^{-\frac{1}{2} (\mathbf{x} - \mathbf{p})^T \boldsymbol{\Sigma}^{-1} (\mathbf{x} - \mathbf{p})} \; .
\end{equation}

The covariance matrix is parameterized by a rotation matrix \(\mathbf{R}\) and a scaling matrix \(\mathbf{S}\) to constrain it to be positive semi-definite:
\begin{equation}
\mathbf{\Sigma} = \mathbf{RSS^TR^T}\
\end{equation}
where \(\mathbf{R}\) and \(\mathbf{S}\) are the matrix representations of the learnable quaternion $\mathbf{r}$ and the 3D scaling $\mathbf{s}$ vectors, respectively.
Then, the 3D Gaussians are projected onto screen space using EWA Splatting \cite{zwicker2002ewa} as follows:
\begin{equation}
\mathbf{\Sigma'} = \mathbf{JW\Sigma W^TJ^T}\ 
\end{equation}
where \(\mathbf{\Sigma'}\)is a projected $2\!\times\!2$ covariance matrix. \( \mathbf{J'} \in \mathbf{R}^{2 \times 3} \) denotes the Jacobian of the affine-approximated projective transformation matrix and \( \mathbf{W} \in \mathbf{R}^{3 \times 3} \) is the viewing transformation. Finally, the pixel-space rendered color \(\mathbf{C}_{r}\) is computed using point-based rendering \cite{3DGS}:
\begin{equation}
\mathbf{C}_{r} = \sum_{i \in \mathrm{N}} c_i a_i \prod_{j=1}^{i-1} (1 - a_j) \; .
\end{equation}
Here, \(\mathrm{N}\) denotes the total number of ordered points. $c_{i}$ represents the per-Gaussian decoded color, and $a_{i}$ represent the result of multiplying the covariance matrix \(\mathbf{\Sigma'}\)with the the per-Gaussian decoded opacity $o_i$, respectively.

\begin{figure}[]
	\centering
    \includegraphics[width=\linewidth]{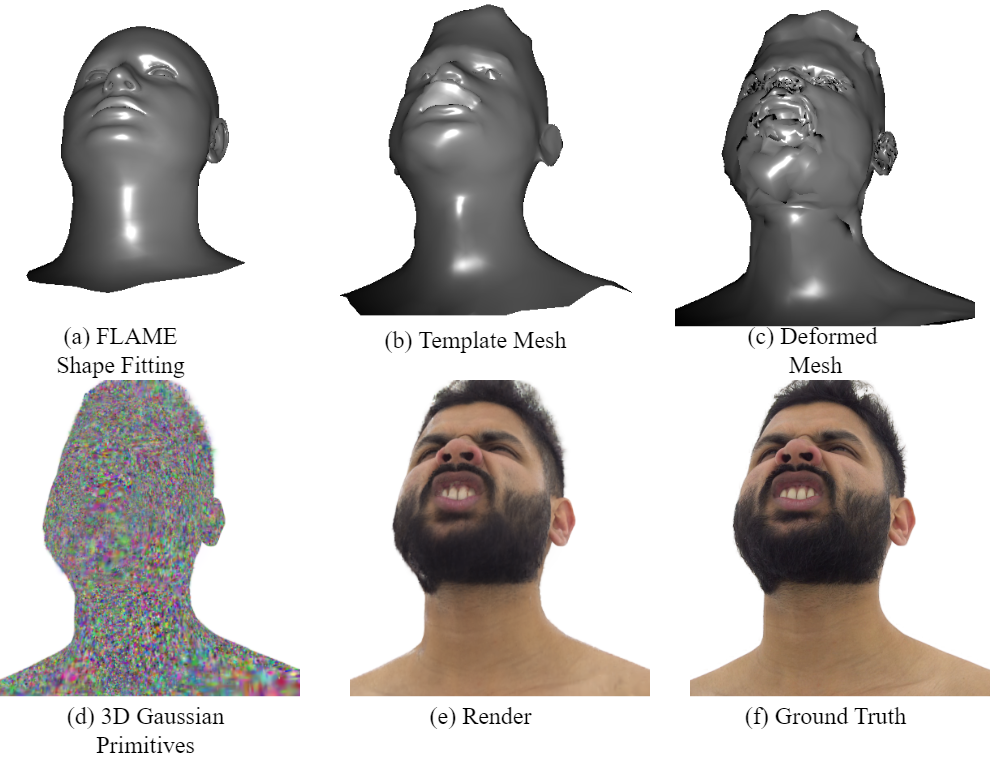}
	\caption{\textbf{Coarse-to-fine Learning.} We refine the FLAME shape-fitted mesh by an initial registration step (b), which results in the template mesh with vertices $v_t$ used in our framework. (c) The template mesh is deformed based on the input and posed to the world space using the global transformation parameters. We initialize the 3D Gaussians on the rigidly translated deformed mesh with vertices and refine their properties through our Coarse-to-Fine framework, resulting in dense, head surface-aligned 3D Gaussians (d). The 3D Gaussians are splatted, resulting in the render (e), which is supervised with ground truth image (f). 
  }
	\label{fig:intermediates}
\end{figure}

\subsection{Encoding}\label{sec:neuraltex}
Unlike 3D Gaussian Splatting \cite{3DGS}, which was originally proposed to model static scenes, we aim to control the 3D Gaussians based on RGB images as input. 
To this end, we introduce an encoder \(\mathbf{E}_{\gamma}\) that takes in \(\mathrm{I}_{\text{d}}=3\) multi-view RGB images as input to encode the facial appearance and the global rigid head pose from varying viewpoints. The encoder is designed to encode the changes in local dynamics like facial expression changes via an animation code \(\mathbf{Z}_{\text{exp}} \in \mathbb{R}^{256}\) and global transformations, separately. Our choice of encoding multi-view RGB images is similar to related works that encode expressions via multi-view RGB images \cite{ziyancompositional,lombardi21_mvp, yang23_travatar,lombardi2018_deep-appear}. Additionally, the encoder outputs rigid transformation parameters, where the rotation matrix \(\mathbf{R}_{\text{rigid}}\) lies in SO(3), and the translation vector \(\mathbf{T}_{\text{rigid}}\) belongs to \(\mathbb{R}^{3}\). The encoder has separate branches for the animation code and the rigid transformation parameters. The rotation output is normalized to ensure it lies on the unit sphere, representing a unit quaternion.  While TRAvatar \cite{tran2019_towards} shares a similar design choice of predicting the global transformation parameters, we find the need in our framework for constraining these parameters via a landmark alignment loss as weak supervision to guide the positioning of the rigid head pose, as discussed in Sec. \ref{subsec:obj_functions}.

\subsection{Coarse-to-fine Learning Framework}\label{sec:c2f}
Our method leverages a coarse-to-fine framework for deforming a template mesh. It first initializes the 3D Gaussians and then refines the positions of the Gaussians. This hierarchical approach is imperative to our high-fidelity results, as discussed in Sec. \ref{sec:ablation}. The coarse step handles the large vertex-level deformations that result from extreme or exaggerated facial expressions. The fine deformation step then refines these changes, allowing for capturing the more fine-scale or intricate details like the teeth. Fig. \ref{fig:intermediates} provides a high-level overview of the coarse-to-fine framework.

\subsubsection{Coarse Step}
Before training, we first register the FLAME-fitted mesh on the first frame of the sequence to multi-view RGB images of the subject. Sec. \ref{sec:implementation_supp} presents the details of this template mesh generation step. 
To deform the template mesh to match the subject's expression closely, we utilize a per-vertex offset prediction network \(\mathbf{D}_{\text{v}}\). This network takes the animation code \(\mathbf{Z}_{\text{exp}}\) as input and decodes per-vertex offsets $\delta_{v}$ on top of a neutral expression template mesh \(\mathbf{M}_{\text{template}}\) with vertices \(\mathbf{v}_{\text{t}}\). This results in deformed vertices \(\mathbf{v}_{\text{d}}\) as follows:
\begin{equation}
\mathbf{v}_{\text{d}}\ = \mathbf{v}_{\text{d}}\ + \delta_{v} \; . \label{eq:1}
\end{equation}
The deformed mesh \(\mathbf{M}_{\text{d}}\) with vertices \(\mathbf{v}_{\text{d}}\) shares the same topology as FLAME \cite{li2017_flame}, with a total of $5023$ vertices. Since the number of vertices is fairly small, we employ a lightweight MLP-based architecture to learn the per-vertex offsets on top of the template vertices \(\mathbf{v}_{\text{t}}\). The vertices of the deformed mesh are then rigidly posed to the world space using the estimated global rigid rotation  \(\mathbf{R}_{\text{rigid}}\) and translation  \(\mathbf{R}_{\text{rigid}}\) as follows: 
\begin{equation}
\mathbf{v}_{\text{p}}\  = \mathbf{R}_{\text{rigid}}(\mathbf{v}_{\text{d}})+\mathbf{T}_{\text{rigid}} \; . \label{eq:2}
\end{equation}
This global transformation results in the posed mesh \(\mathbf{M}_{\text{p}}\), defined by vertices \(\mathbf{v}_{\text{p}}\).

\subsubsection{3D Gaussian Initialization}\label{sec:3dIntilaization}
We initialize \(\mathrm{N}_{\text{G}}\) 3D Gaussians on the posed mesh \(\mathbf{M}_{\text{p}}\). To do so, we first uniformly sample on the UV space of the template mesh at a resolution of  $512 \times 512$. This allows for sampling \(\mathrm{N}_{\text{G}}=\mathrm{N}^{2}_{\text{g}}\) 3D Gaussians in 2D space, where \(\mathrm{N}_{\text{g}}\) corresponds to the spatial resolution in the \(\mathbf{x}\) and \(\mathbf{y}\) direction of the UV map. Once sampled, each Gaussian is assigned a position \(\mathbf{p}_{\text{in}}\), rotation \(\mathbf{r}_{\text{in}}\), and scale \(\mathbf{s}_{\text{in}}\). The position \(\mathbf{p}_{\text{in}}\) is the interpolated vertex position on the deformed mesh, computed using barycentric interpolation on the UV map. For the first frame, the scales are initialized according to a distance transform-based technique akin to 3D Gaussian Splatting~\cite{3DGS}, which serves as the initial scale reference \(\mathbf{s}_{\text{in}}\). The initial rotations \(\mathbf{r}_{\text{in}}\) are set to zero. We apply a binary mask that filters out the samples that do not lie on the UV parameterization of the template mesh.

\subsubsection{Fine Step}\label{sec:deltadecoders}
A 2D CNN-based decoder \(\mathbf{D}_{\text{m}}\) is used to decode the animation code \(\mathbf{Z}_{\text{exp}}\) into deltas $(\delta_p, \delta_r, \delta_s)$, which represent the offsets in positions, rotations, and scales with respect to the initialized Gaussian primitives. Thus, we predict the positions $\mathbf{p} = \mathbf{p}_{\text{in}} + \delta_p$, rotations $\mathbf{r} = \mathbf{r}_{\text{in}} + \delta_r$, and scales $\mathbf{s} = \mathbf{s}_{\text{in}} + \delta_s$ for the 3D Gaussians. We use a 2D CNN to generate the deltas in a grid of size \(\mathrm{N}_{\text{G}}\times\mathrm{N}_{\text{G}}\times10\). This grid corresponds to the primitives sampled in the UV map. The position offsets $\delta_p$ represent the fine-level deformation of facial structures showing fine-grained features, such as facial hair and mouth interior, as these offsets allow for 3D Gaussians to be placed in these regions. 
% Adjust the value to reduce the 
\subsection{Gaussian Opacity and Color Decoding}\label{sec:rgbadecoders}
To achieve real-time and high-fidelity renderings, it is crucial to decode the opacity and appearance information of each 3D Gaussian efficiently. This is achieved through computationally efficient decoders that translate the animation code into per-Gaussian properties, capturing both the opacity and RGB color information. Our approach leverages 2D CNNs for efficient decoding of both RGB and opacity values. This design choice is similar to several concurrent works \cite{zhu2023ash,saito2024rgca, jiang2024uv}.

\subsubsection{Opacity Decoder}The animation code \(\mathbf{Z}_{\text{exp}}\) is decoded into a 2D opacity map \(\mathbf{O}\) of size \(\mathrm{N}_{\text{G}}\times\mathrm{N}_{\text{G}}\times1\) using \(\mathbf{D}_{\text{o}}\). Each value on the grid represents the per-Gaussian opacity \(\mathbf{o}_{\text{i}}\).

\subsubsection{RGB Decoder} 
The animation code \(\mathbf{Z}_{\text{exp}}\), along with the object-centric view direction \(\mathbf{v}\), is decoded into a 2D color map \(\mathrm{N}_{\text{G}}\times\mathrm{N}_{\text{G}}\times3\) using \(\mathbf{D}_{\text{RGB}}\). Each value on the decoded grid represents the per-Gaussian RGB color \(\mathbf{c}_{\text{i}}\). Note that the view direction \(\mathbf{v}\) is introduced to represent view-dependent appearance effects.

\subsection{Objective Function}
\label{subsec:obj_functions}

Given the above representation, we learn the encoder \(\mathbf{E}_{\text{gamma}}\) and the decoders \(\mathbf{D}_{\text{v}}\), \(\mathbf{D}_{\text{o}}\), \(\mathbf{D}_{\text{RGB}}\), and \(\mathbf{D}_{\text{m}}\) in an end-to-end manner using multi-view image supervision. 
For this, we optimize for the following objective function:
\begin{align}
\mathbf{L}\ = & \ \lambda_1 \mathbf{L}_{\text{1}}\ + \lambda_2 \mathbf{L}_{\text{SSIM}}\ + \lambda_3 \mathbf{L}_{\text{geo}}\ + \lambda_4 \mathbf{L}_{\text{perc}}\ \notag \\
    & + \lambda_5 \mathbf{L}_{\text{temp}}\ + \lambda_6 \mathbf{L}_{\text{lmk}}\ + \lambda_7 \mathbf{L}_{\text{reg}}\ .
\end{align}

Here, \(\mathbf{L}_{\text{1}}\) denotes the difference between the rendered RGB image \(\mathbf{I}_{\text{r}}\) and ground truth RGB image \(\mathbf{I}_{\text{g}}\). The perceptual distance, \(\mathbf{L}_{\text{perc}}\), is based on a pretrained VGG network \cite{simonyan2015deep}.  Similar to \cite{cao2022_authentic}, we take 4 random patches of size 500$\times$500 pixels from the ground truth and rendered image pair to compute the perceptual distance. \(\mathbf{L}_{\text{SSIM}}\) denotes the structural similarity measure \cite{ssim}. To supervise the deformed mesh \(\mathbf{M}_{\text{d}}\) with vertices \(\mathbf{v}_{\text{d}}\), we employ an \(\mathbf{L}_{\text{2}}\) loss between the deformed vertices \(\mathbf{v}_{\text{d}}\) and the tracked vertices \(\mathbf{v}_{\text{s}}\) as follows:
the space before the 
\begin{equation}
\mathbf{L}_{\text{geo}}(\mathbf{v}_{\text{d}}, \mathbf{v}_{\text{s}}) = \frac{1}{\mathrm{N}_{\text{v}}} \sum_{i=1}^{\mathbf{N}_{\text{v}}} (\mathbf{v}_{\text{d}}[i] - \mathbf{v}_{\text{s}}[i])^2 \; ,
\end{equation}
where \(\mathrm{N}_{\text{v}}\) represents the number of vertices.
\(\mathbf{L}_{\text{geo}}\) allows the vertices to move freely towards more deformable regions like the mouth interior while remaining close to the tracked vertices \(\mathbf{v}_{\text{t}}\). This acts as a soft prior to preserve the structure of the face. We further use a spatial regularization term
\(\mathbf{L}_{\text{reg}} \) that ensures stable training. \(\mathbf{L}_{\text{reg}} \) is implemented as the summation of L1 penalties for scales $\mathbf{s}$, Gaussian mean position offsets \( \delta p \), and Gaussian rotations offsets \( \delta r \):
the space before the 
\begin{equation}
\mathbf{L}_{\text{reg}} = \frac{1}{\mathbf{N}_{\text{G}}} \sum_{i=1}^{\mathrm{N}_{\text{G}}} \left( |\mathbf{s}_i| + |\delta p_i| + |\delta r_i| \right)
\end{equation}
Here, \(\mathrm{N}_{\text{G}} \) represents the total number of Gaussians. In addition, we employ a landmark constraint $\mathbf{L}_{\text{lmk}}$ to supervise the global transformation. This constraint ensures that the rigid landmarks of the face align with the rigid landmarks of the rigidly tracked head pose mesh \(\mathbf{M}_{\text{r}} \), which is at neutral expression. 
 % Adjust the value to reduce the space before the 
\begin{equation}
\mathbf{L}_{\text{lmk}} =\frac{1}{\mathrm{N}_{\text{L}}}  \sum_{i=1}^{\mathbf{N}_{\text{L}}} \left| \mathbf{M}_{\text{p,i}} - \mathbf{M}_{\text{r,i}} \right| 
\end{equation}
Here, \(\mathrm{N}_{\text{L}} \) is the total number of rigid landmarks. 
\(\mathbf{M}_{\text{p,i}} \) represents the \(i\)-th rigid landmark of the posed mesh \(\mathbf{M}_{\text{p}} \) and 
\(\mathbf{M}_{\text{r,i}} \) denotes the corresponding \(i\)-th rigid landmark of the rigidly tracked mesh \(\mathbf{M}_{\text{r}} \) at a neutral expression.
Finally, we employ a temporal smoothness loss $\mathbf{L}_{\text{temp}}$ between two consecutive frames at time $t$ and $t+1$, defined on the Gaussian offsets predicted by the deltas decoder \(\mathbf{D}_{\text{m}}\) as follows:
\begin{align}
\mathbf{L}_{\text{temp}} = \frac{1}{\mathrm{N}_{\text{G}}} \sum_{i=1}^{\mathrm{N}_{\text{G}}} \Big( & \left| \delta p_i^{(t+1)} - \delta p_i^{(t)} \right| 
    + \left| \delta \mathbf{s}_{\text{i}}^{(t+1)} - \delta \mathbf{s}_{\text{i}}^{(t)} \right| \notag \\
    & + \left| \delta r_i^{(t+1)} - \delta r_i^{(t)} \right| \Big) 
\end{align}

In our experiments, we set $\lambda_1=0.8$, $\lambda_2=0.2$, $\lambda_3=0.1$, $\lambda_4=0.01$, $\lambda_5=0.1$, $\lambda_6=0.8$, and $\lambda_7=0.1$. 

\subsection{Implementation Details}
\label{subsec:implementation_details}
 We train our method for $300$k iterations on a single NVIDIA $A40$ GPU, which takes around $24$ hours. We adopt the Adam optimizer \cite{KingBa15} for estimating the optimal parameters of our learning-based framework. We set the learning rate to $1e{-4}$ for \(\mathbf{D}_{\text{v}}\). The Opacity decoder \(\mathbf{D}_{\text{o}}\) and RGB decoder \(\mathbf{D}_{\text{RGB}}\) are trained with a learning rate of $6e{-4}$, while the image encoder \(\mathbf{E}_{\gamma}\)  is trained with a learning rate of $5e{-4}$. The delta decoder \(\mathbf{D}_{\text{m}}\) is trained with a learning rate of $1e{-4}$. We use \(\mathrm{N}_{\text{L}}=4\) rigid landmarks ($2$ for each eye) to supervise the landmark alignment. In our framework, the view direction $\mathbf{v}$ is calculated as the mean position of the subject's head defined by the posed mesh \(\mathbf{M}_{\text{p}}\) with vertices \(\mathbf{v}_{\text{p}}\) subtracted by the camera viewpoint center.

\section{Experiments} 

\begin{figure}[!ht]
\centering
\includegraphics[width=0.48\textwidth]{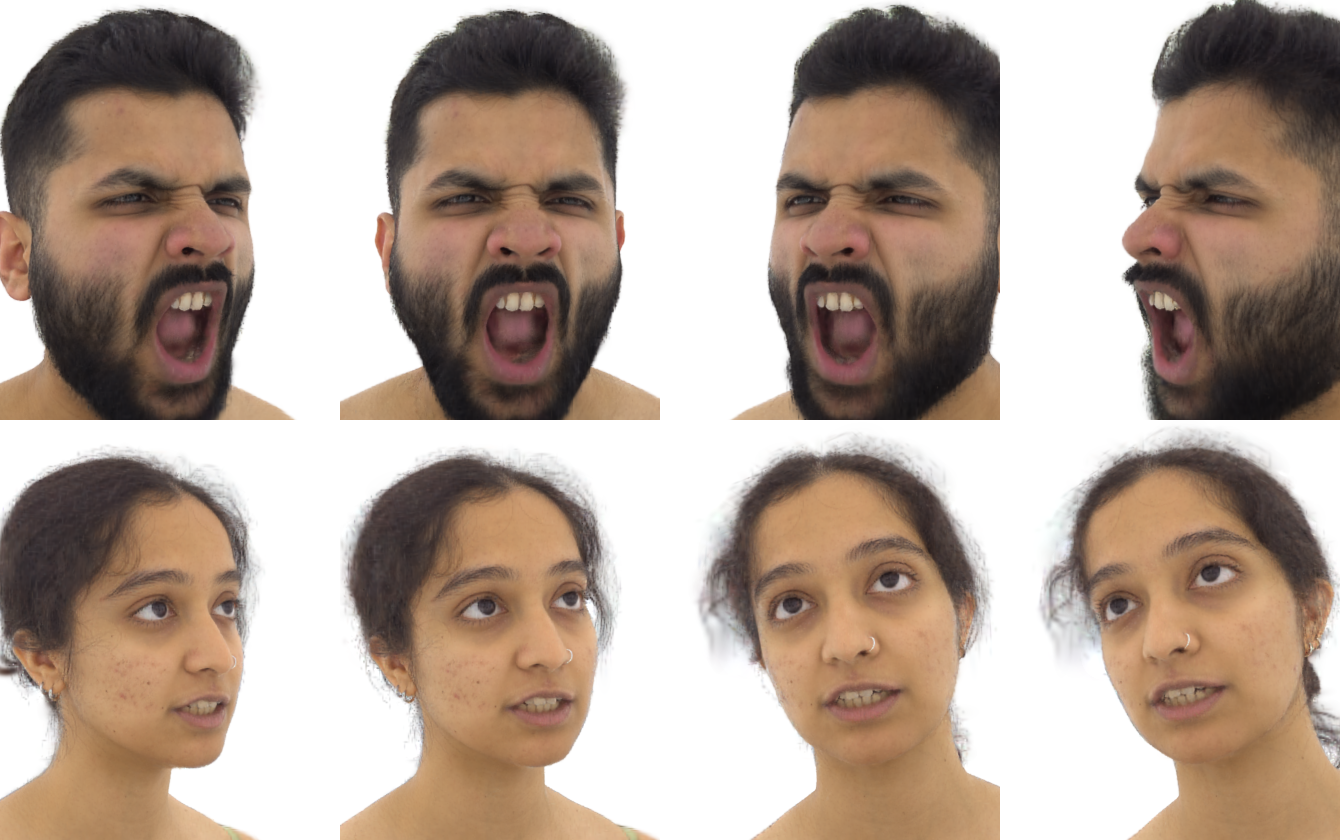}
\caption{\textbf{Qualitative Results.} Novel view synthesis at virtual camera paths for different subjects. Our method excels at representing fine details, such as facial hair and fine scalp hair strands.}
\label{fig:qualitative_results-nvs}
\end{figure}

Our method produces photorealistic moving heads under novel viewpoints and with unseen expressions. 
This section discusses the various experiments we have performed to examine our method.
We show results on multiple participants recorded in a multi-view facial performance capture rig. Unless stated otherwise, all our experiments used for evaluations are trained with $23$ cameras, sparsely distributed over the entire head. We hold out $1$ camera for evaluation. Furthermore, the default resolution used in our experiments is $960\times540$, unless stated otherwise. We show qualitative and quantitative comparisons on $4$ subjects performing challenging facial expressions, which includes $1$ subject from the NeRSemble dataset \cite{kirschstein2023nersemble}. We hold out $300$ frames for both quantitative and qualitative evaluation unless stated otherwise. 
We first show qualitative results for a wide variety of sequences (Sec. ~\ref{sec:subjective}). Through quantitative and qualitative analysis, we highlight the important design choices of our method (Sec. ~\ref{sec:ablation}) and perform comparisons against related work, namely MVP~\cite{lombardi21_mvp}, GaussianAvatars \cite{qian24_gaussian-avatars}, and HQ3DAvatar~\cite{teotia2023hq3davatar}.
For image sequence results, please refer to the supplemental video.
\subsection{Qualitative Results}\label{sec:subjective}
\begin{figure}[!t]
\centering
\includegraphics[width=\linewidth]{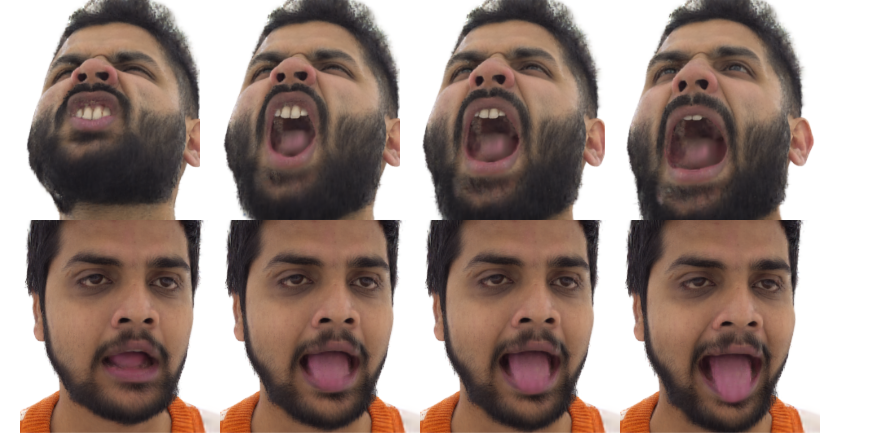}
\caption{\textbf{Qualitative Results.} Expression synthesis from a novel camera viewpoint. Our approach can synthesize challenging expressions and motions, such as the tongue sticking out.}
\label{fig:qualitative_results-exp_changes}
\end{figure}
\begin{figure}[ht]
\centering
\includegraphics[width=0.48\textwidth]{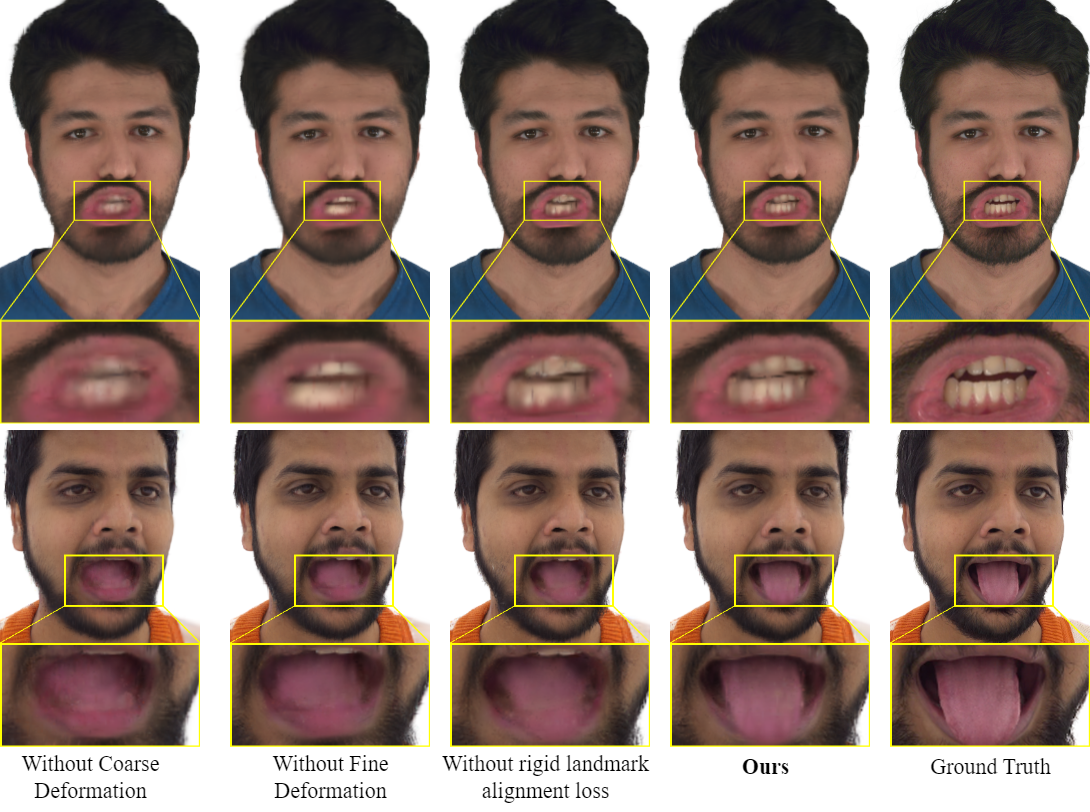}
\caption{\textbf{Ablation Study.} \textit{Left to right}: Without coarse deformations, without fine deformations, without rigid landmark alignment loss  \(\mathbf{L}_{\text{lmk}}\), ours, and Ground Truth. The highlighted regions show expression alignment in the mouth region. We observe that our design choices result in high-quality details in the mouth interior. Note: Best viewed if zoomed in or in the supplementary video. }
\label{fig:ablations_components}
\end{figure}

 Fig.~\ref{fig:qualitative_results-nvs} shows turntable renderings generated by our approach. Fig.~\ref{fig:qualitative_results-exp_changes} shows novel expressions synthesized for $2$ different subjects at a holdout camera viewpoint. Our results show high-quality view synthesis, as well as novel expression synthesis.  We also show an application that demonstrates that our model can work with inputs from a parametric face tracker \cite{li2017_flame} in the supplementary video.
\subsection{Ablative Analysis}\label{sec:ablation}
\begin{figure}[ht]
\centering
\includegraphics[width=0.48\textwidth]{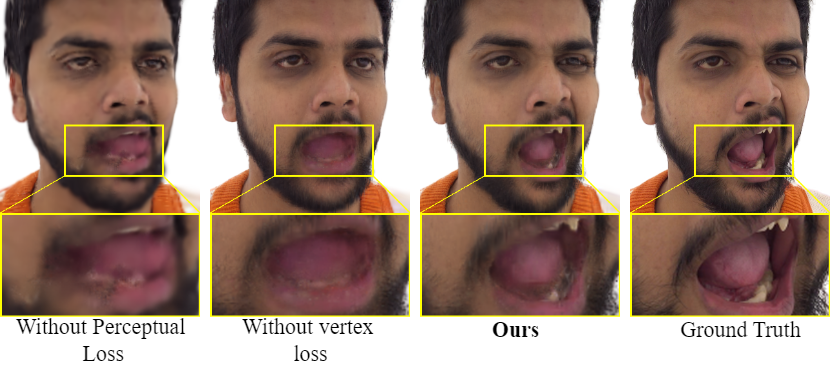}
\caption{\textbf{Ablation Study.} \textit{Left to right}: Without perceptual loss,without vertex loss  \(\mathbf{L}_{\text{geo}}\), and ours. The highlighted regions show expression alignment in the mouth region. Note that using a perceptual loss term leads to sharper facial hair and tongue reconstructions, while  \(\mathbf{L}_{\text{geo}}\) helps with detailed expression alignment.}
\label{fig:ablations_loss}
\end{figure}
\begin{figure}[!ht]
\centering
\includegraphics[width=0.46\textwidth]{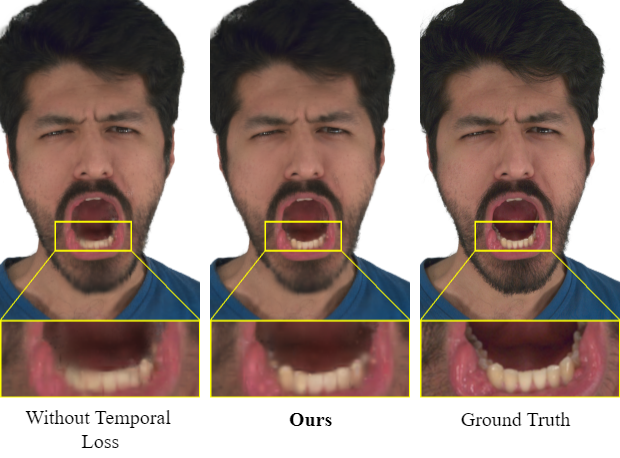}
\caption{\textbf{Ablation Study.} \textit{Left to right}: Without  \(\mathbf{L}_{\text{temp}}\), ours, and Ground Truth.  \(\mathbf{L}_{\text{temp}}\) helps reproduce details in structures undergoing large motion changes like teeth.}
\label{fig:ablations_temporal}
\end{figure}
\begin{figure}[!ht]
\centering
\includegraphics[width=0.48\textwidth]{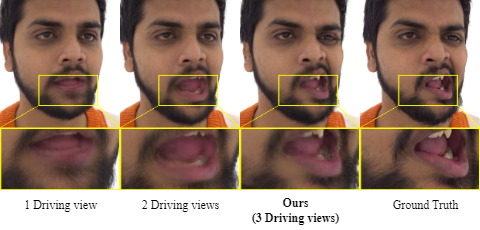}
\caption{\textbf{Ablation Study.} \textit{Left to right}: $1$ input driving view, $2$ input driving views, ours ($3$ input driving views), and Ground Truth. We observe that our choice of driving the avatar with an image-triplet leads to overall better expression alignment.}
\label{fig:dv}
\end{figure}
\begin{figure}[!ht]
\centering
\includegraphics[width=0.48\textwidth]{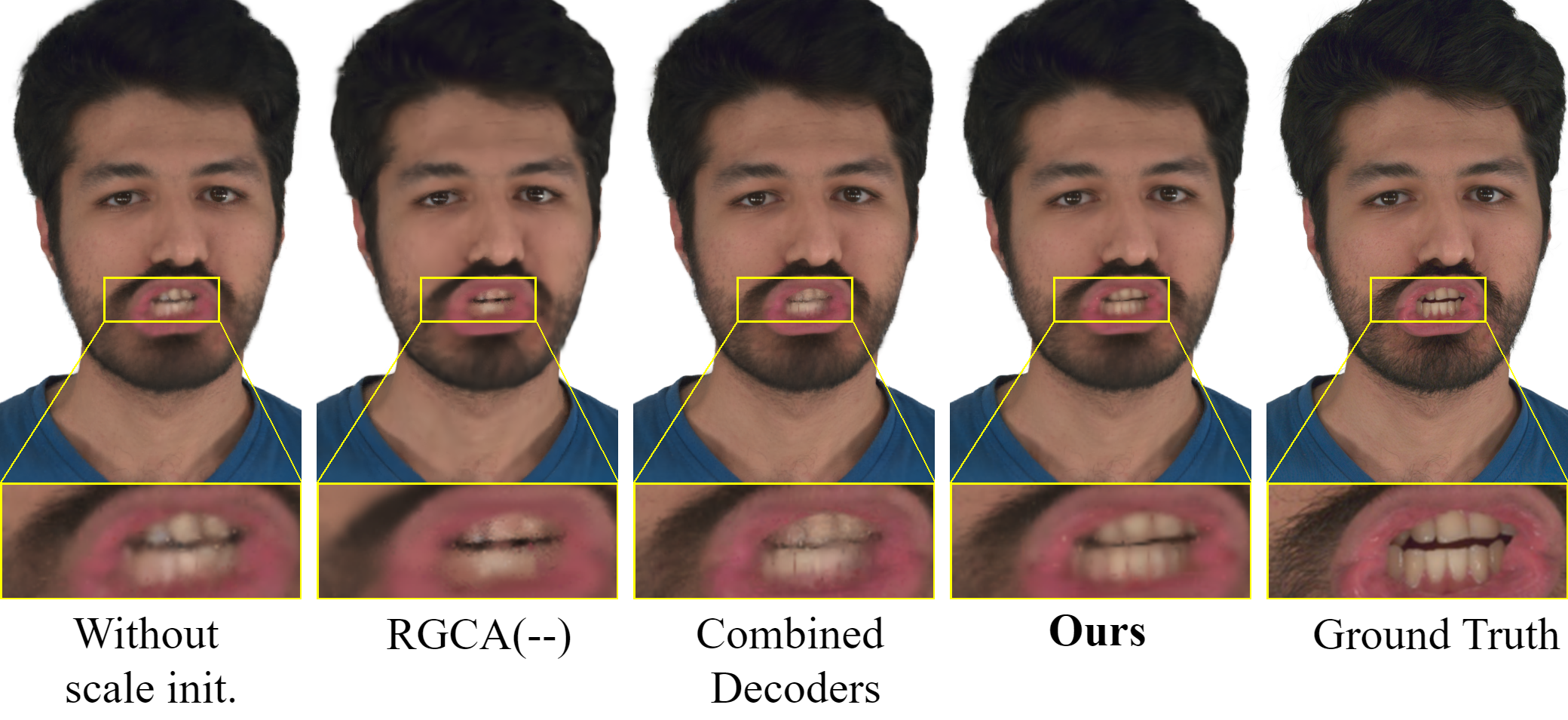}
\caption{\textbf{Ablation Study.} \textit{Left to Right}: Without scale initialization, RGCA (- -), using combined decoders for 3D Gaussian geometry properties, ours, and Ground Truth. Our design choice generates sharper results in deformable regions like the mouth.}
\label{fig:rgcamm}
\end{figure}
\begin{table}[!t]
\centering
\caption{\textbf{Analysis of our core design choices}. We outperform the baseline approaches in quantitative metrics.}
\label{tab:ablation_study_1}
\begin{tabular}{|l|c|c|c|c|}
\hline
Metrics & No Coarse Def. & No Fine Def. & No \(\mathbf{L}_{\text{lmk}}\) & \textbf{Ours} \\ \hline
PSNR ↑ & 32.95 & 32.91 & 33.08 & \textbf{33.91} \\ \hline
L1 ↓ & 7.44 & 7.56 & 7.41 & \textbf{4.93} \\ \hline
SSIM ↑ & 0.80 & 0.80 & 0.80 & \textbf{0.85} \\ \hline
LPIPS ↓ & 0.13 & 0.13 & \textbf{0.11} & \textbf{0.11} \\ \hline
\end{tabular}
\end{table}
\begin{table}[!ht]
\centering
\caption{\textbf{Ablation Study}: Quantitative evaluation of temporal loss term  \(\mathbf{L}_{\text{temp}}\), perceptual loss term  \(\mathbf{L}_{\text{perc}}\), and vertex loss  \(\mathbf{L}_{\text{geo}}\). Our method attains the best numerical performance on almost all metrics (see bold numbers).}
\resizebox{0.48\textwidth}{!}{
\begin{tabular}{|l|c|c|c|c|}
\hline
Metrics & Without  \(\mathbf{L}_{\text{temp}}\) & Without  \(\mathbf{L}_{\text{perc}}\) & Without  \(\mathbf{L}_{\text{geo}}\) & \textbf{Ours} \\ \hline
PSNR ↑ & 33.64 & 33.08 & 32.98 & \textbf{33.91} \\ \hline
L1 ↓ & \textbf{4.88} & 7.37 & 7.35 & 4.93 \\ \hline
SSIM ↑ & \textbf{0.85} & 0.81 & 0.80 & \textbf{0.85} \\ \hline
LPIPS ↓ & \textbf{0.11} & 0.15 & \textbf{0.11} & \textbf{0.11} \\ \hline
\end{tabular}
}

\label{tab:ablation_study}

\end{table}

\begin{table}[!t]
\centering
\caption{\textbf{Analysis of the number of input driving views}. We observe that driving with more input views increasingly improves head avatar synthesis.}
\label{tab:driving_views}
\small{
\begin{tabular}{|c|c|c|c|c|}
\hline
Metrics & $1$ input view & $2$ input views & \textbf{Ours ($3$ input views)}  \\
\hline
PSNR $\uparrow$  & 30.94 & 31.78 & \textbf{33.11} \\
\hline
L1 $\downarrow$ & 15.52 & 9.38 & \textbf{5.58} \\
\hline
SSIM $\uparrow$ & 0.67 & 0.75 & \textbf{0.83}  \\
\hline

\end{tabular}
}
\end{table}
\begin{table}[!ht]
\centering
\caption{\textbf{Ablation Study}: Quantitative evaluation with combined decoders for predicting Gaussian properties, RGCA (- -), and without using scale initialization. Our method attains the best numerical performance (see bold numbers).}
\resizebox{0.48\textwidth}{!}{
\begin{tabular}{|l|c|c|c|c|}
\hline
Metrics & Combined Decoders & RGCA (- -) & No Scale Init. & \textbf{Ours} \\ \hline
PSNR ↑ & 34.25 & 34.11 & 33.80 & \textbf{34.79} \\ \hline
L1 ↓ & 4.80 & 4.95 & 5.29 & \textbf{3.91} \\ \hline
SSIM ↑ & 0.86 & 0.86 & 0.85 & \textbf{0.87} \\ \hline
LPIPS ↓ & \textbf{0.10} & 0.12 & 0.11 & \textbf{0.10} \\ \hline
\end{tabular}
}

\label{tab:ablation_study_new}
\end{table}
We ablate the main components as well as loss terms employed in our approach. The ablation study in Fig.~\ref{fig:ablations_components} demonstrates the importance of our core components. Without coarse deformation, the model struggles to recover the structure in the mouth interior. Without fine deformation, the model still lacks fine-grained detail of facial structures, such as the teeth. Without landmark alignment loss \(\mathbf{L}_{\text{lmk}}\), the model misaligns expressions in 
the mouth. Our model successfully captures the details in the mouth interior at a high quality. Fig. \ref{fig:ablations_loss} shows the impact of the perceptual loss term \(\mathbf{L}_{\text{perc}}\) and the per-vertex loss term \(\mathbf{L}_{\text{geo}}\). The absence of \(\mathbf{L}_{\text{perc}}\) results in overall blurry face renderings. \(\mathbf{L}_{\text{geo}}\) helps with expression alignment in regions undergoing large deformations, e.g., the mouth. Fig. \ref{fig:ablations_temporal} highlights the impact of temporal loss \(\mathbf{L}_{\text{temp}}\). We find that \(\mathbf{L}_{\text{temp}}\) helps preserve fine-scale details of structures undergoing motion across time, such as the teeth. Table~\ref{tab:ablation_study_1} shows the average image-based quality assessment metric scores. We use PSNR, SSIM \cite{wang2004_image-quality}, L1 distance, and LPIPS distance \cite{zhang2018_unreasonable-effectiveness} as the image-based quality assessment metrics. The scores are averaged over the holdout frames for $2$ subjects. Numerical results show that our design choices are well-validated. Tab. \ref{tab:ablation_study} shows a quantitative evaluation for different design choices, including the use of the temporal loss term  \(\mathbf{L}_{\text{temp}}\), the perceptual loss term  \(\mathbf{L}_{\text{perc}}\), and the vertex loss term  \(\mathbf{L}_{\text{geo}}\). We observe that our final method, and our method without  \(\mathbf{L}_{\text{temp}}\) performs on par on quantitative metrics as its effects are more apparent during abrupt facial deformations. Quantitative evaluations focusing solely on the mouth region show improvements in the L1 metric (7.27 with the temporal loss and 7.41 without). To quantify the temporal stability for  \(\mathbf{L}_{\text{temp}}\), we run the JOD metric \cite{JOD} with and without  \(\mathbf{L}_{\text{temp}}\) in place, which yields superior results when using the proposed loss term (7.10 with the temporal loss vs 7.04 without), where a higher number suggests better quality. Fig. \ref{fig:dv} provides a qualitative comparison on the design choice of using 3 input driving views on Subject $1$'s data. We observe that our design choice of using an image-triplet to encode the facial expressions leads to overall better results. Fig. \ref{fig:rgcamm} shows a qualitative comparison for different baselines. They include no initialization for Gaussian scales, our implementation of RGCA decoders~\cite{saito24_relightable-gaussian},
combined decoders for Gaussian opacity and offset values, and ours. We remark that, unlike our method, RGCA utilizes combined decoders for Gaussian opacity and offset values. We observe that our results reproduce sharper details around the mouth interior than those obtained by the other approaches. Tab. \ref{tab:ablation_study_new} compares the same baselines against our approach on Subject $\mathbf{S_{NeRSemble_1}}$ and supports our qualitative results.

\begin{figure*}[ht]
\centering
\includegraphics[width=0.8\textwidth]{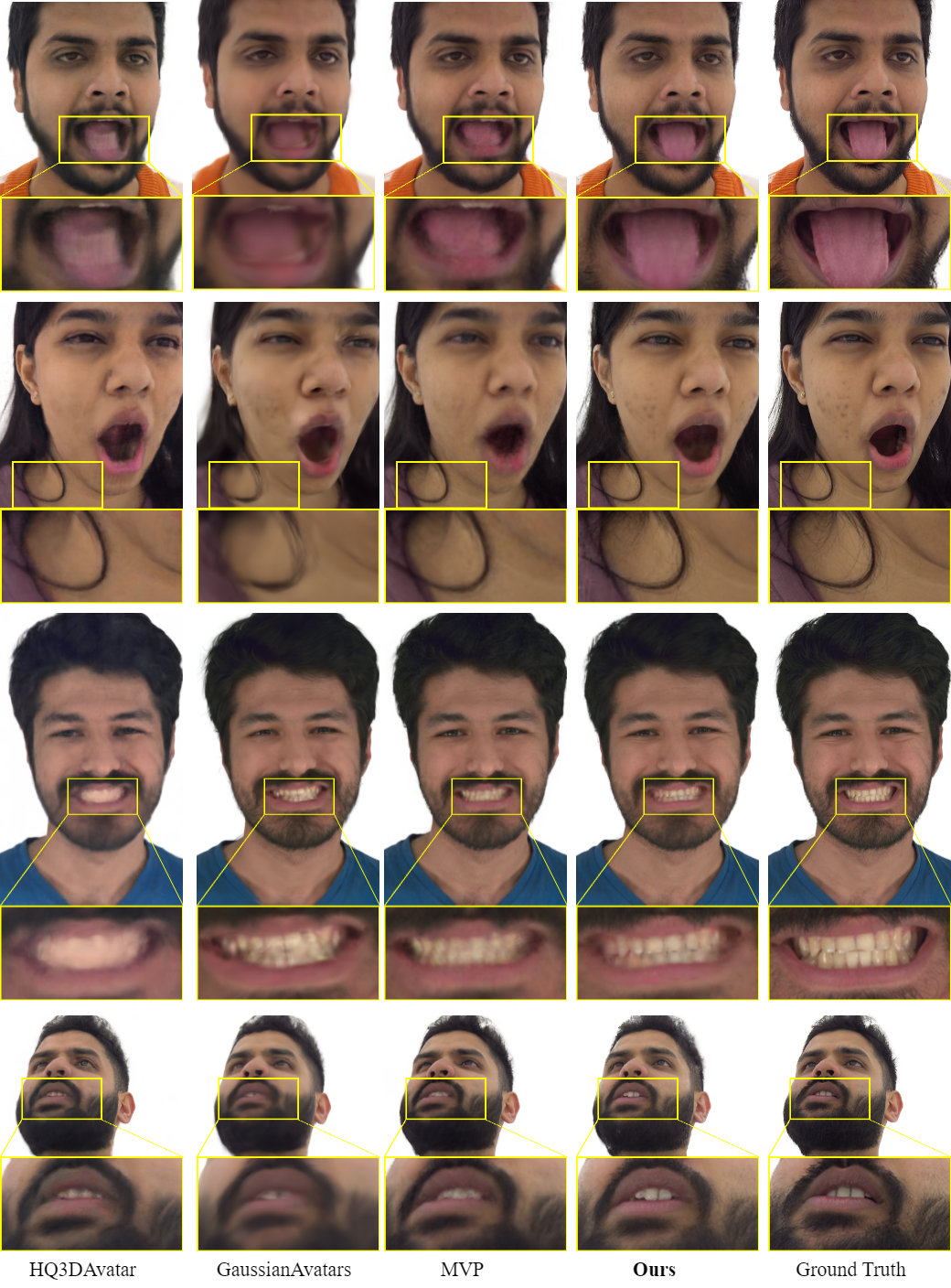}
\caption{\textbf{Qualitative comparisons} with the related methods. \textit{Top to bottom}: Subject $1$, Subject $2$, Subject $\mathbf{S_{NeRSemble_1}}$, and Subject $3$. \textit{Left to right}: HQ3DAvatar \cite{teotia2023hq3davatar}, GaussianAvatrars \cite{qian24_gaussian-avatars}, MVP~\cite{lombardi21_mvp},  and Ours. Note that yellow boxes highlight the detail reproduction with respect to the ground truth. Our method produces crisper details in the mouth interior and hair.}
\label{fig:comparison_stota}
\end{figure*}
\subsection{Comparisons}
\begin{figure}[!ht]
\centering
\includegraphics[width=0.44\textwidth]{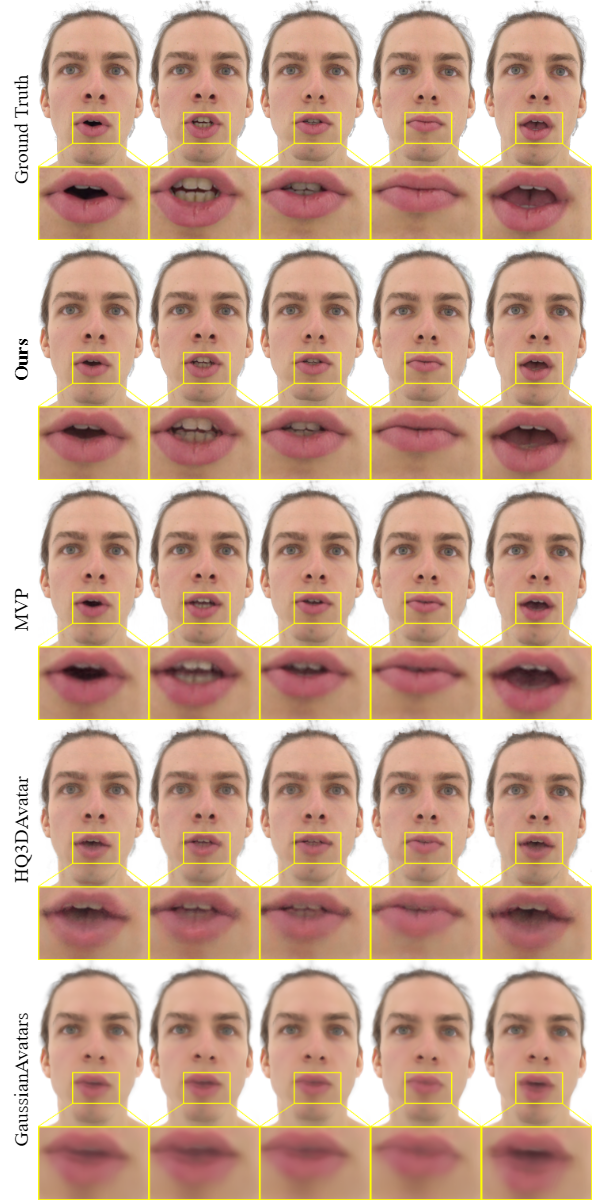}
\caption{\textbf{Qualitative comparison on a speech sequence}. Our method results in overall better mouth-region synthesis.}
\label{fig:audio}
\end{figure}

\begin{table}[!t]
\centering
\caption{\textbf{Quantitative comparison with related approaches}. We outperform the related methods numerically on several image-quality metrics (see bold text).}
\label{tab:quantitative_results-ave_error-sota}
\small{
\begin{tabular}{|c|c|c|c|c|}
\hline
Metrics & HQ3DAvatar & GaussianAvatars & MVP & \textbf{Ours} \\
\hline
PSNR $\uparrow$  & 30.78 & 32.22 & 32.50 & \textbf{32.89} \\
\hline
L1 $\downarrow$ & 13.68 & 8.90 & 9.03 & \textbf{7.33} \\
\hline
SSIM $\uparrow$ & 0.77 & 0.79 & 0.78 & \textbf{0.81} \\
\hline
LPIPS $\downarrow$ & 0.14 & 0.15 & 0.15 & \textbf{0.11} \\
\hline
\end{tabular}
}
\end{table}

We compare our real-time multi-view dynamic head generation method against three multi-view techniques: HQ3DAvatar \cite{teotia2023hq3davatar}, GaussianAvatars \cite{qian24_gaussian-avatars}, and Mixture of Volumetric Primitives (MVP) \cite{lombardi21_mvp}. Notably, GaussianAvatars \cite{qian24_gaussian-avatars}, MVP \cite{lombardi21_mvp}, and our method operate in real time. HQ3DAvatar employs an implicitly learned canonical radiance field, while GaussianAvatars registers 3D Gaussians on a tracked mesh. MVP uses ray-marching through cubic volumetric primitives on a template mesh. MVP \cite{lombardi21_mvp} and our approach can work with RGB images as input, whereas Gaussian Avatars \cite{qian24_gaussian-avatars} is designed to work with a parametric face model. Fig.~\ref{fig:comparison_stota} presents qualitative reconstruction comparisons. HQ3DAvatar \cite{teotia2023hq3davatar}, GaussianAvatars \cite{qian24_gaussian-avatars}, and MVP \cite{lombardi21_mvp} struggle with topological changes, fine-scale hair details, dental features, and beard textures, often producing blurry and less defined results. In contrast, our method captures these details more accurately and sharply, as evidenced by the sharper representation of the tongue (top row), fine hair-strand structure (second row), clear dental features (third row), and distinct beard textures (bottom row).
Table~\ref{tab:quantitative_results-ave_error-sota} shows the quantitative performance of our approach against MVP~\cite{lombardi21_mvp} and HQ3DAvatar~\cite{teotia2023hq3davatar} averaged over holdout frames for $4$ subjects. We observe that for the image-quality assessment metrics,
our method shows the best numerical performance. \\
Fig. \ref{fig:audio} shows a qualitative comparison of our method against the different baselines for a speech sequence trained on a subject speaking phonetically balanced sentences. We observe that our approach outperforms baseline methods in terms of image synthesis quality, especially for the mouth region.

\section{Limitations and Future Work}
\begin{figure}[!ht]
\centering
\includegraphics[width=0.46\textwidth]{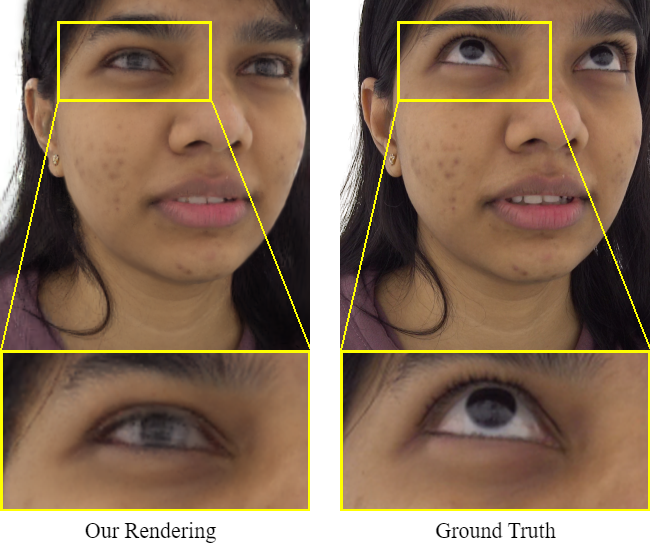}
\caption{\textbf{Failure Case.} \textit{Left to right}: Our rendering and Ground Truth. Our model produces ghosting artifacts for subtle movements in the eye region.}
\label{fig:limit}
\end{figure}

While our method clearly sets the state of the art for human head avatar rendering in terms of speed and photorealism, it still has certain limitations. 
First, the image encoder takes the entire face region as input, which prevents independent animation of local face regions, such as the cheeks. It also struggles with subtle movements of the eyes, as highlighted in Fig. \ref{fig:limit}. Future approaches could integrate localized animation models for fine-grained control. Second, the choice of sampling could be made more efficient.
Future directions could involve sampling a low number of Gaussians and learning their initial positioning on the deformed template surface rather than uniformly sampling a dense number of Gaussians.
Third, we rely on a multi-view camera rig during training. 
In the future, we plan to investigate learning photoreal head avatars using more lightweight setups.
For driving the avatar from outside of the multiview capture setup, an augmentation strategy similar to \cite{clothing} could help reduce the domain gap between the capture and driving environments by color-augmenting the driving images. At the same time, future research should investigate robustness to lighting changes or even truthful disentanglement of lighting and material properties. Additionally, audio-based appearance synthesis is also a promising future direction that our current work can enable.
\section{Conclusion} \label{sec:conclusion}

We have presented a novel end-to-end framework to render high-quality moving human heads with large deformations at real-time speed. 
By representing the head model in a coarse-to-fine manner, our method first deforms a template mesh based on an animation code extracted from input images.
This has been shown to be effective in capturing coarse-level facial deformations, such as wide mouth opening and large head pose changes.
We then sample the 3D Gaussians, which are parameterized by the animation code, on the deformed mesh via 2D UV space to further refine the deformation and the geometry into detailed facial expressions, such as lip rolling and fine-grained teeth shape.
Utilizing a 3D splatting rasterization technique enables not only accelerating end-to-end training into $24$ hours, but also real-time inference in up to 75 FPS.

With the global transformations as a learnable parameter, our method does not require an accurate, explicit global rigid pose, as opposed to competing radiance field rendering approaches.
We demonstrate that the above significantly helps our model outperform state-of-the-art methods visually and numerically using various metrics.
We also present a comprehensive ablative analysis of the proposed loss terms and component-level design decisions.
Finally, we show the robustness of our method with some challenging scenes with highly dynamic head poses and deformations.

\bibliographystyle{ACM-Reference-Format}
\bibliography{references}
\appendix

\section{Implementation details} \label{sec:method_supp}

\begin{figure}[!ht]
\centering
\includegraphics[width=0.46\textwidth]{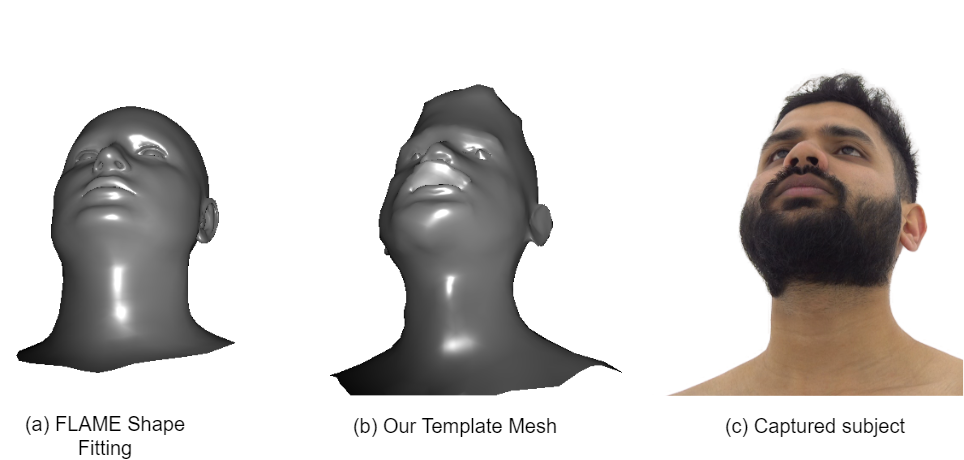}
\caption{\textit{Left to right}: We deform the initial FLAME shape mesh (a) to fit the photometric data, resulting in a deformed template mesh used for training our model (b). (c) Shows the captured identity.}
\label{fig:template_update}
\end{figure}
We provide further details of our model implementation in the following section.
\subsection{Template registration}\label{sec:implementation_supp}

\begin{figure*}[!ht]
\centering
\includegraphics[width=0.86\textwidth]{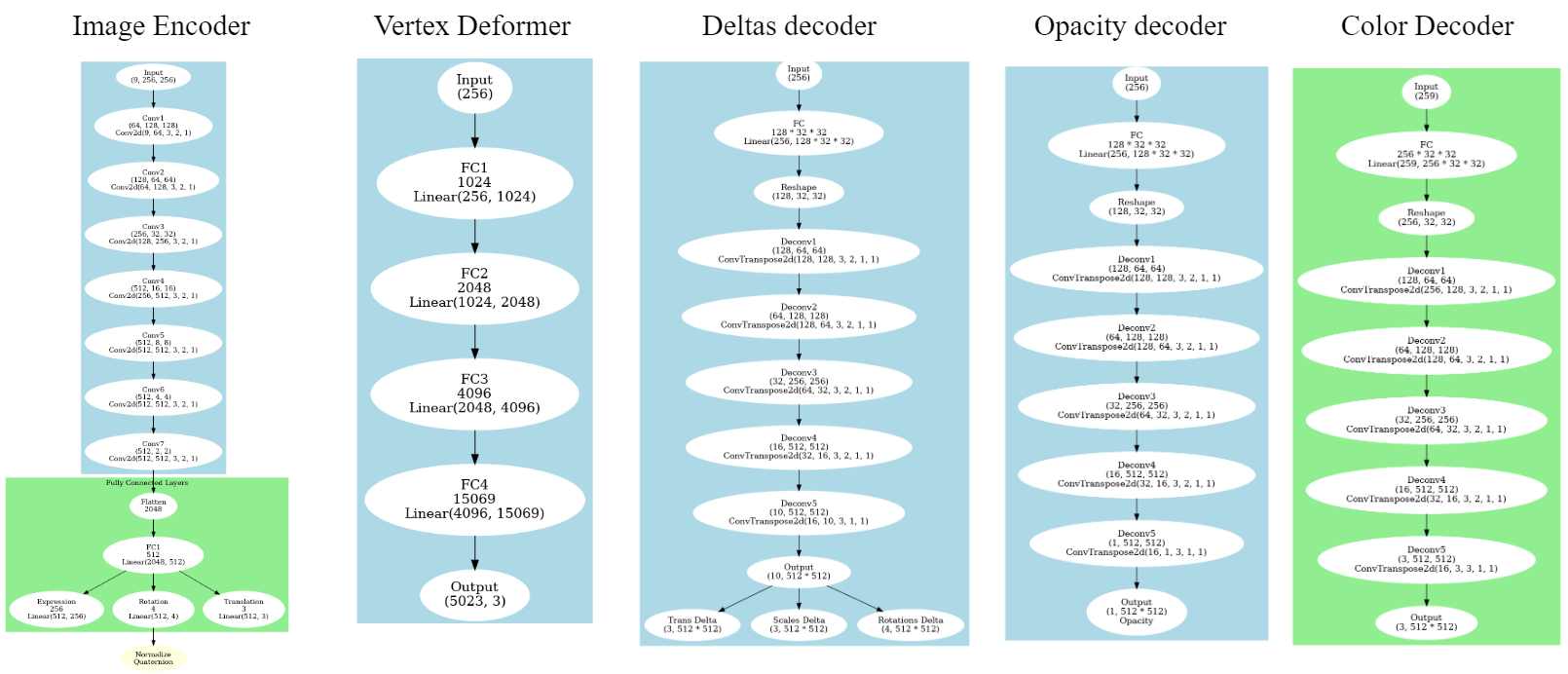}
\caption{Architecture details of the different components employed in our approach.}
\label{fig:network}
\end{figure*}

We deform the FLAME \cite{li2017_flame} template to match the photometric data of multiview RGB images by using our method as a renderer. This step is highlighted in Fig.~\ref{fig:template_update}. We use the FLAME \cite{li2017_flame} template as the base template mesh and learn per-vertex offsets for a static frame. In addition to the photometric loss terms used in our approach, we use a landmark-based loss term and a Laplacian-smoothness term to obtain a smooth template mesh.\\
The details of our network architecture, such as network operators and feature dimensions, are illustrated in Fig. \ref{fig:network}.
\begin{figure}[!ht]
\centering
\includegraphics[width=0.48\textwidth]{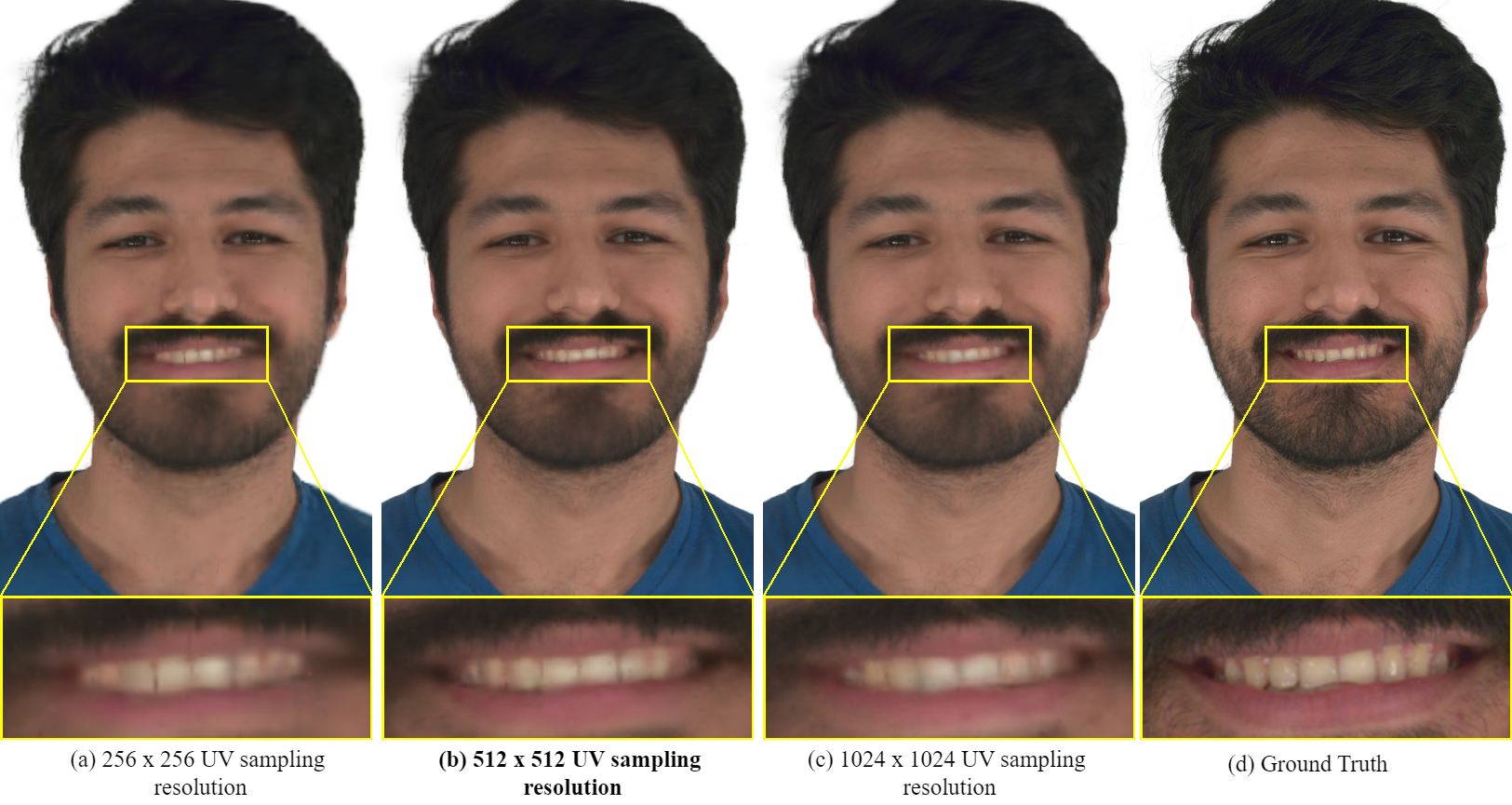}
\caption{Qualitative results: Choice of sampling resolution. We highlight that a sampling resolution of 512 $\times$ 512 leads to comparative results to $1024\times1024$, i.e., the maximum sampling resolution. A 256 $\times$ 256 sampling resolution leads to blurry details inside the mouth. Rendering with sampling at full $1024\times1024$ is twice as slow as ours. }
\label{fig:numsamps2}
\end{figure}

\subsection{Baselines for Comparisons}\label{sec:baselines_supp}
In Sec. 4.3, we compare our approach against HQ3DAvatar \cite{teotia2023hq3davatar}, MVP \cite{lombardi21_mvp}, and GaussianAvatars \cite{qian24_gaussian-avatars}. As both MVP and HQ3DAvatar can take an image as input, we provide both methods with the same multiview triplet input as ours. In addition, GaussianAvatars \cite{lombardi21_mvp} and MVP \cite{lombardi21_mvp} are provided with the same template mesh used by our approach. For comparison against RGCA \cite{saito24_relightable-gaussian}, we only implemented the decoding framework of RGCA \cite{saito24_relightable-gaussian}, which uses shared decoders for Gaussian geometry property prediction, tracked expression meshes for initializing the Gaussians, and no scale initialization.

\section{Experiments} 
\begin{table}[!ht]
\centering
\caption{\textbf{Runtime comparison for various rendering components}. The table lists the components needed for rendering the final image and their associated computational times (in seconds). The total runtime is $0.0132$ seconds or ~75 FPS.}
\resizebox{0.48\textwidth}{!}{
\begin{tabular}{|c|c|c|c|c|c|c|c|}
\hline
\text{Component} & $E_{\gamma}$& $D_{v}$ & $D_{RGB}$ &  $D_{o}$ & $D_m$ & Sampling & Rasterization  \\
\hline
\text{Time} & 0.0012 & 0.0002 & 0.0008 & 0.0008  & 0.0007 &0.0001 &0.0094\\
\hline
\end{tabular}
}

\label{tab:component_runtime}
\end{table}

Table \ref{tab:component_runtime} shows the runtime breakdown of our method. The runtime is evaluated on the holdout frames of $3$ subjects. Rasterization is the most expensive operation, which still runs over 100 fps. Fig. \ref{fig:numsamps2} shows a qualitative comparison that initializes 3D Gaussians with different UV map sampling resolutions.

\end{document}